# LXPER Index 2.0: Improving Text Readability Assessment for L2 English Learners in South Korea

**Bruce W. Lee[1,2]**
Dep. of Computer & Information Science[1]
University of Pennsylvania
PA, USA
brucelws@seas.upenn.edu

**Jason Hyung-Jong Lee[2]**
Research & Development Center[2]
LXPER, Inc.
Seoul, South Korea
jasonlee@lxper.com

## Abstract

Developing a text readability assessment model specifically for texts in a foreign English Language Training (ELT) curriculum has never had much attention in the field of Natural Language Processing. Hence, most developed models show extremely low accuracy for L2 English texts, up to the point where not many even serve as a fair comparison. In this paper, we investigate a text readability assessment model for L2 English learners in Korea. In accordance, we improve and expand the Text Corpus of the Korean ELT curriculum (CoKEC-text). Each text is labeled with its target grade level. We train our model with CoKEC-text and significantly improve the accuracy of readability assessment for texts in the Korean ELT curriculum.

## 1 Introduction

Text readability assessment has been an important field of research since the 1940s. However, most research focused on the native audience in English speaking countries (Benjamin, 2012). In China, Japan, and Korea, many high and middle school students attend English language schools, in addition to their regular school classes. English subject plays an important role in the educational systems of the three countries (Mckay, 2002).

Despite the importance put in English education, the previous text readability assessment models have not been in active use in the three countries. This is due to the poor performance of traditional readability assessment models on L2 texts. We believe there is an immediate need for the development of an improved text readability assessment method for use in L2 education around the world. In this research, we put a specific focus on L2 English learners in South Korea. But our methodology is applicable to other ELT (English Language Training) curricula.

Many traditional readability assessment models are linear regression models with a small number of linguistic features, consisting of the generic features of a text like total words, total sentences, and total syllables (Kincaid et al., 1975). Such features are effective predictors of a text's readability, but more curriculum-specific features are required for L2 text readability assessments. The key distinction between native readability assessment and L2 readability assessment is that L2 students rigorously follow the specific national ELT curriculum. Unlike native students who learn English from a variety of sources, most L2 students have limited exposure to English. In this research, we reduce the average assessment error by implementing some curriculum-specific features.

The contributions of this paper are: (1) we utilize and expand CoKEC-text, one of the few graded corpora with texts from an actual L2 curriculum; (2) we investigate novel linguistic features that were rarely tested on an L2 corpus; (3) we evaluate our model against other readability models, show significantly improved accuracy, and prove that "grades" are better modeled using logistic regression, not linear regression.

## 2 Related Work

Research efforts in developing automated text readability assessment models for L2 students only emerged in the 2000s (Xia et al., 2016). Heilman et al. (2007) showed that grammatical features and lexical features play particularly important roles in L2 text readability prediction. Meanwhile, Vajjala and Meurers (2014) showed that the additional use of lexical features could significantly improve L2 readability assessment. Feng et al. (2010) also reported the importance of lexical features in general (for L1 speakers of English) text readability assessment.

However, the common limitation of the previous research in L2 readability assessment was the training corpus annotated with the grade levels for L1 readers of English. Our results, which we obtain from training our model using CoKEC-text, introduces the possibility that lexical features are not as important as the previous researchers reported. In addition, we also show that a considerably accurate text readability model can built even with a small data set if the model is optimized and the corpus is well-labeled.

## 3 Corpus

Since our goal is to improve the accuracy in text readability assessment of L2 texts, our ideal corpus has to fully consist of L2 texts from a non-native ELT curriculum. The base corpus that we use is CoKEC-text (Lee and Lee, 2020), which is a collection of 2760 unique grade-labeled texts that are officially administered by the Korean Ministry of Education (MOE). Similar texts are also used in the National Assessment of Educational Achievement, College Scholastic Ability Test, and MOEapproved middle school textbooks in Korea.

However, as shown in Table 1, the number of texts in the original CoKEC-text is heavily skewed to higher grades (K10 ∼ K12.5) than in lower grades (K7 ∼ K9). Such a disparity can affect the accuracy of our regression results and can become troublesome in predicting the lower grade texts' readability. Thus, we decided to collect about 900 more texts from Korean MOE-approved middle school textbooks and use them to create an expanded version of CoKEC-text.

In addition, we found some K7 ∼ K9 texts that are only partially English. They contained ASCII Korean Characters (often explanations of difficult English words by the author). These count as a token (or possibly, even a sentence from case to case) in the NLTK parsing process but provide no meaningful linguistic properties. This can produce miscalculations of the "average of x" (e.g., the average number of words per sentence) features that we discuss in Section 4. We manually went through every text to make sure that clean data is used for model training. Our final training corpus consists of 3700 original L2 texts from K7 ∼ K12.5. K12.5 grade texts are from CSAT, which is a college entrance exam for Korean universities. In general, the Korean grades K7 to K12 are for middle and high school students of ages 13 to 19.

| Difficulty Labels | Original | Expanded |
|---|---|---|
| 12.5 | 691 | 691 |
| 12 | 590 | 601 |
| 11 | 596 | 602 |
| 10 | 571 | 580 |
| 9 | 80 | 313 |
| 8 | 215 | 302 |
| 7 | 17 | 305 |

Table 1: Number of texts in two corpus versions

## 4 Selecting features

We now describe the 35 linguistic features we studied. Table 2 contains a list of the features with a shortcode name used throughout this paper. The list is divided into five parts: traditional features, POS-based features, entity density features, lexical chain features, and word difficulty features.

### 4.1 Traditional Features

We first implemented some traditional features from the popular Flesch-Kincaid model: aWPS (average number of Words per Sentence), aSPW (average number of Syllables per Word), and P3T (words with more than 3 syllables per Text). These are one of the earliest linguistic features studied in text readability prediction, but they prove to be still useful in recent studies (Feng et al., 2010).

### 4.2 POS-based Features

A number of researchers commonly reported that POS-based features are effective in text readability prediction. In particular, Peterson and Ostendorf (2009) investigated the following features: aNP, aNN, aVP, aAdj, aSBr, aPP, nNP, nNN, nVP, nAdj, aSBr, nPP. Lee and Lee (2020) proved that these features are highly correlated with the difficulty of L2 texts. However, their dataset mostly consisted of K10, K11, K12 texts, and their evaluation was conducted only on K9 ∼ K12. Thus, there exists a possibility that the result was heavily influenced by higher grade L2 texts. We evaluate these features again with our expanded version of CoKEC-text.

### 4.3 Entity Density Features

We implement entity density features in an attempt to account for the difficulty in comprehending conceptual information in texts.

Such information is often introduced by entities, or more specifically, general nouns and named entities.

The density of entities introduced in a text relates to the working memory burden, which has an increasing trend in a positive correlation with the age of the reader. Our main task is to develop a model that would be particularly useful to L2 student groups, and accurately classify the given texts to the respective student grade level. Hence, we believe that these entity density features are great predictors. Some of these features were never tested on an L2 corpus, and the results we obtain are novel.

### 4.4 Lexical Chain Features

We believe that the accuracy of L2 text readability can be improved by incorporating lexical chain features as well. Since L2 readers have limited exposure to English compared to native readers, we hypothesize that L2 readers work harder in connecting several entities and recognizing the semantic relationship. Entities that form these semantic relations are connected throughout the text in the form of lexical chains. However, in Table 3 we observe that lexical chain features are weakly correlated to target grade levels of L2 texts in Korea.

### 4.5 Word Difficulty Features

Native English readers learn vocabulary from a variety of sources. On the other hand, most L2 students learn new English words step by step, following the respective national ELT curriculum. Hence, implementing curriculum specific features related to vocabularies can be particularly useful in predicting the text difficulty for L2 students.

We use CoKEC-word to identify the difficulty of words (Lee and Lee, 2020). The word corpus is a classification of 30608 words in 6 levels. It only consists of the words that previously appeared in the Korean ELT curriculum. We focused on the vocabularies in levels B, C, D, E, and F. This covers vocabularies from K5 to college level.

### 4.6 Parsing and Counting Modules

We used a combination of spaCy (popular opensource library for NLP) (Honnibal and Montani, 2017), NLTK (NLP toolkit for Python) (Bird et al., 2009), Gensim (famous for topic modeling and FastText model) (Řehůřek and Sojka, 2010), and the Berkeley Neural Parser (constituency parser) (Kitaev and Klein, 2018) to parse and count the features described in this section.

| Code | Feature Description |
|---|---|
| aWS | average number of Words per sent |
| aSPW | avg num of Syllables per word |
| P3T | % of words with $\geq$ to 3 syl |
| nWD | total number of Words per Doc |
| aNP | avg num of Noun Phrases per sent |
| aNN | avg num of proper nouns per sent |
| aVP | avg num of Verb Phrases per sent |
| aAdj | avg num of Adjectives per sent |
| aSBr | avg num of Subord. Clauses per sent |
| aPP | avg num of Prepos. Phrases per sent |
| nNP | total num of Noun Phrases per sent |
| nNN | total num of proper nouns per sent |
| nVP | total num of Verb Phrases per sent |
| nAdj | total number of Adjectives per sent |
| nSBr | total num of Subord. Clauses per sent |
| nPP | total num of Prepos. Phrases per sent |
| PND | % of named entities per doc |
| PNS | % of named entities per sent |
| nUE | total number of Unique Entities |
| aEM | avg num of Entity Mentions per sent |
| aUE | avg num of Unique Entities per sent |
| nLC | total num of Lexical Chains |
| aLCW | avg num of Lexical Chains per word |
| aLCS | avg num of Lex Chains per noun sent |
| aLCN | avg num of Lex Chains per noun phrase |
| nBw | total num of lev B (K5-8) words |
| aBw | avg num of lev B words per word |
| nCw | total num of lev C (K8-9) words |
| aCw | avg num of lev C words per word |
| nDw | total num of lev D (K9-11) words |
| aDw | avg num of lev D words per word |
| nEw | total num of lev E (K11-12) words |
| aEw | avg num of lev E words per word |
| nEw | total num of lev F (college) words |
| aFw | avg num of lev F words per word |

Table 2: Number of texts in two corpus versions

To keep operation simple, only NLTK and Berkeley Neural Parser were used in the previous version of LXPER Index. However, our further investigation show that certain tasks are performed at much higher accuracy by complementary libraries. For example, spaCy showed the highest accuracy at recognizing a sentence, and Gensim improved the lexical chaining process.

### 4.7 Selecting Features

We computed the Pearson correlation value of each feature and checked if it was significant enough (correlation > 0.07) in predicting the target grade level of a text. The "Cor" column in Table 3 lists the correlation value of each feature. We ordered the list in decreasing correlation values. Next, we removed the features that are highly correlated ("Paired?" column). The "Include?" column in Table 3 summarizes the final features.

| Code | Cor | Sig? | Paired? | Include? |
|------|-----|------|---------|----------|
| nDw  | 0.532 | Y | Y | Y |
| aWS  | 0.512 | Y | N | Y |
| aSPW | 0.499 | Y | N | Y |
| aDw  | 0.487 | Y | Y | N |
| nBw  | 0.454 | Y | Y | Y |
| aNP  | 0.446 | Y | N | Y |
| P3T  | 0.444 | Y | N | Y |
| aNN  | 0.434 | Y | N | Y |
| aPP  | 0.423 | Y | N | Y |
| nPP  | 0.417 | Y | N | Y |
| nCw  | 0.402 | Y | Y | Y |
| nEw  | 0.399 | Y | Y | Y |
| nAdj | 0.394 | Y | N | Y |
| aAdj | 0.378 | Y | N | Y |
| nNN  | 0.376 | Y | N | Y |
| aVP  | 0.323 | Y | N | Y |
| nWD  | 0.321 | Y | N | Y |
| nNP  | 0.308 | Y | N | Y |
| aSBr | 0.298 | Y | N | Y |
| aCw  | 0.289 | Y | Y | N |
| aBw  | 0.274 | Y | Y | N |
| nSBr | 0.221 | Y | N | Y |
| aEw  | 0.221 | Y | Y | Y |
| nLC  | 0.212 | Y | N | Y |
| PND  | 0.201 | Y | N | Y |
| nEw  | 0.195 | Y | Y | Y |
| PNS  | 0.174 | Y | N | Y |
| aLCW | 0.154 | Y | N | Y |
| nVP  | 0.126 | Y | N | Y |
| aLCN | 0.0995 | Y | N | Y |
| aFw  | 0.0976 | Y | Y | N |
| aLCS | 0.0913 | Y | N | Y |
| aUE  | 0.0884 | Y | N | Y |
| aEM  | 0.0792 | N | N | N |
| nUE  | 0.00833 | N | N | N |

Table 3: Selecting features

## 5 Readability Assessment

We built a logistic regression model and trained it with the new expanded version of CoKEC-text to complete our assessment tool; our model is programmed in Python. To evaluate the new model's effectiveness for L2 students in Korea, we prepared a separate test corpus.

| Type | F-K | D-C | LX 1.0 | LX 2.0 |
|------|-----|-----|--------|--------|
| K7   | 4.89 | 5.38 | 9.21 | 7.3 |
| K8   | 5.44 | 5.02 | 9.43 | 8.45 |
| K9   | 5.78 | 5.53 | 9.86 | 9.04 |
| K10  | 10.6 | 7.95 | 10.9 | 10.5 |
| K11  | 9.66 | 7.57 | 11.4 | 11.3 |
| K12  | 9.21 | 7.31 | 11.5 | 11.6 |
| Avg Er. (in K) | 2.10 | 3.04 | 1.05 | 0.34 |

Table 4: Final results

The first part (K10 ~ K12) of our test corpus is from the official mock tests that were used by KICE (Korea Institute of Curriculum & Evaluation) to assess the educational achievement of high school students from 2017 to 2020. There are 270 texts in the first part of our test corpus (K10: 90 texts, K11: 90 texts, K12: 90 texts). The second part of our corpus is from the government-approved middle school textbooks (K7: 90 texts, K8: 90 texts, K9: 90 texts).

We collected the texts from two sources to test how our readability assessment model performs on different types of texts. Ideally, our results should show a continuous increase from K7 to K12 texts. Our target average assessment error is below 0.5 grade level. Table 4 summarizes our results. We compare our LXPER Index 2.0 (LX 2.0) to traditionally popular models like Flesch-Kincaid (F-K) (Kincaid et al., 1975), Dale-Chall (D-C)

(Dale and Chall, 1949), and the previous LXPER Index 1.0 (LX 1.0) (Lee and Lee, 2020).

We also wanted to compare our model to the more recently developed models, but we could not find any suitable L2 readability index. We attempted comparison with Lexile Score and CohMetrix L2 Readability Score (Crossley et al., 2008). However, the models had a completely different grading scale and did not show a consistently increasing trend with grades. This was also reported in our previous research (Lee and Lee, 2020).

## 6 Conclusion

In this research, we introduced LXPER Index 2.0, a readability assessment tool that incorporates traditional, POS, entity density, lexical chain, and word difficulty features. Then, we trained the model on our own expanded version of CoKEC-text. We obtained a continuously increasing output for L2 texts from K7 to K12. In addition, we achieved our initial target average accuracy error of less than 0.5 grade levels, which is more accurate than any L2 text readability prediction model we are aware of.

The improvements we report in this paper are largely due to two changes: 1. the CoKEC expansion and 2. the use of a logistic regression model. The contribution from the corpus is quite obvious in that our model could now learn more about the lower grades (K7 ∼ K9). However, the contribution from the change of the regression model is something that we should put more thought into. But it seems evident that the "grades" classification task is better modeled with a logistic regression model. A possible explanation could be that the difficulty of a text does not linearly correlate with the target grades.

Even though we wanted to test our model on other East Asia L2 ELT curricula, like Japan and China, we could not implement due to the lack of openly-available corpus in the countries. The novelty of the LXPER Index model is that it focuses on in-curriculum text readability analysis, possibly even with a small data set of less than 4000 texts. Thus, applying the model will fail to give meaningful outcomes without a pre-processed and labeled corpus like CoKEC. Thus, the application of a similar model on those countries would first require foundation research on constructing corpora.